\documentclass[letterpaper]{article} % DO NOT CHANGE THIS
\usepackage{aaai2027} % DO NOT CHANGE THIS
\usepackage[hyphens]{url} % DO NOT CHANGE THIS
\usepackage{graphicx} % DO NOT CHANGE THIS
\usepackage{natbib} % DO NOT CHANGE THIS AND DO NOT ADD OPTIONS
\usepackage{caption} % DO NOT CHANGE THIS AND DO NOT ADD OPTIONS
\usepackage{booktabs}
\usepackage{amsmath}
\usepackage{amssymb}

\usepackage{multirow}
\usepackage[table]{xcolor}
\usepackage{colortbl}
\definecolor{HeaderBlue}{RGB}{41, 84, 144}     % deep blue header
\definecolor{OursRow}{RGB}{232, 240, 253}      % light blue highlight for dLoop
\definecolor{BlockGray}{RGB}{247, 247, 249}    % alternating benchmark blocks
\definecolor{BestRed}{RGB}{176, 32, 32}        % best accuracy
\definecolor{BestBlue}{RGB}{23, 92, 176}       % best TPF

\newcommand{\bestacc}[1]{\textcolor{BestRed}{\(\mathbf{#1}\)}}
\newcommand{\besttpf}[1]{\textcolor{BestBlue}{\(\mathbf{#1}\)}}

\title{dLoop: Looped Diffusion Language Models via Recurrent Latent Refinement}

\author{
    Handa Li\textsuperscript{\rm 1}\equalcontrib,
    Pengxiang Li\textsuperscript{\rm 2}\equalcontrib,
    Mingyu Cao\textsuperscript{\rm 1},
    Xilu Wang\textsuperscript{\rm 1},
    Li Shen\textsuperscript{\rm 3},
    Da Li\textsuperscript{\rm 4},
    Lu Yin\textsuperscript{\rm 5}
}
\affiliations{
    \textsuperscript{\rm 1}University of Surrey\\
    \textsuperscript{\rm 2}The Hong Kong Polytechnic University\\
    \textsuperscript{\rm 3}Sun Yat-sen University\\
    \textsuperscript{\rm 4}University of York\\
    \textsuperscript{\rm 5}Shenzhen University of Advanced Technology
}

\nocopyright

\begin{document}

\maketitle

\begin{abstract}
Diffusion language models (DLMs) support parallel generation, yet they often require many denoising steps because only a fraction of token predictions are sufficiently reliable to be committed at each step. Existing acceleration methods primarily optimize confidence learning or commitment policies. However, these approaches operate on the outputs of a single denoising forward pass, in which information from the prompt, previously committed tokens, and uncommitted positions may not yet be fully integrated. Consequently, some predictions remain under-refined when the commitment decision is made, and modifying the commitment policy alone cannot improve their underlying representations. We introduce dLoop, which restructures a pretrained diffusion language model into a Prelude–Loop–Coda architecture and recurrently applies a parameter-shared intermediate block to the same partially masked state. Rather than advancing the denoising process, each additional loop provides further latent computation, allowing information distributed across uncommitted positions to continue interacting and become more fully integrated before commitment. Experiments on GSM8K, MATH, MBPP, and HumanEval show that dLoop achieves the highest or tied-highest accuracy on every benchmark. With tokens per forward pass (TPF) adjusted for loop computation, it outperforms Fast-dLLM in both accuracy and TPF on all four benchmarks and dParallel on three, and is more accurate than d3LLM on all four, while d3LLM attains higher TPF; no compared method surpasses dLoop in both. Mechanistic analyses show that extra loops mainly help on slow, unstable trajectories and that updates at uncommitted positions drive changes in token decisions and later generation. Pre-commitment latent computation thus makes predictions more reliable, offering a route to DLM acceleration distinct from forcing confidence through the training loss.
\end{abstract}

\section{Introduction}

Diffusion language models (DLMs) progressively resolve masked tokens through
iterative denoising, providing a parallel alternative to left-to-right
autoregressive decoding \citep{nie2025llada}. By predicting multiple positions
in one forward pass, DLMs have the potential for parallel generation.
High-quality generation, however, often requires many denoising steps because
only a subset of predictions is reliable enough to commit at each step.
Enabling more predictions to become reliable in fewer steps without
sacrificing generation quality is therefore central to efficient DLM
inference.

Existing acceleration methods primarily improve token commitment through
training or decoding. dParallel uses certainty-forcing distillation to make
correct masked-token predictions reach high confidence earlier
\citep{chen2025dparallel}, and d3LLM retains this certainty-forcing loss while
adding pseudo-trajectory distillation and entropy-guided multi-block decoding
\citep{qian2026d3llm}. Fast-dLLM combines confidence-aware parallel decoding
with approximate KV caching to increase parallel commitment and reduce the
cost of repeated denoising \citep{wu2025fastdllm}. Other approaches use
confidence estimation, local commitment windows, or decoding schedules to
determine when and which predictions should be finalized. These methods mainly
improve when predictions become commit-ready, when tokens are committed, or
the cost of repeated denoising.

Nevertheless, some predictions may remain immature when the decoder must make
a commitment decision. The decoder can commit or defer them, but cannot
further improve the representations from which they were formed. When many
positions remain masked, information relevant to a prediction is distributed
across the prompt, committed positions, and other uncommitted representations
\citep{he2026latent}, and may not be fully propagated in one forward pass.
Rather than only optimizing how existing predictions are committed, we ask a
complementary question: can a model continue computing before irreversible
commitment, allowing immature predictions to be further refined into more
reliable token distributions?

Recurrent-depth models provide a natural mechanism for this purpose.
Repeatedly executing parameter-shared Transformer layers before producing an
output allows hidden states to evolve in continuous latent space, expanding
test-time computation without generating additional external tokens
\citep{geiping2025recurrent,mcleish2025retrofitted,prairie2026parcae}. This
process can be viewed as latent computation: the model iteratively processes
its internal representations before mapping the resulting state to a token
prediction. Recent work applies recurrence to masked diffusion models and
demonstrates benefits from increased loop depth \citep{lee2026loopmdm}.
However, how such pre-commitment computation changes uncommitted
representations and subsequently affects token commitment remains
insufficiently understood.

Motivated by this perspective, we introduce dLoop, which performs recurrent
latent computation within each denoising step before commitment. The model
keeps the current partially masked state unchanged and repeatedly executes a
parameter-shared intermediate block. Extra loops neither reveal new tokens nor
advance to a new external denoising step; instead, they allow the hidden
representations of uncommitted positions to continue evolving and interacting.
Only after these extra passes does the model produce the predictions used for
commitment. Increasing recurrent depth is therefore not an end in itself, but
a mechanism for further refining immature predictions before they become
fixed context.

Starting from a pretrained LLaDA checkpoint, we instantiate dLoop using a
Prelude--Loop--Coda architecture and adapt it through supervised fine-tuning.
The Prelude encodes the current partially masked sequence, the parameter-shared
Loop Block performs one or more additional passes over the same state, and the
Coda maps the updated representations to token distributions. No token is
committed and the model does not advance to a new denoising state during an
extra loop. A confidence-aware decoder then converts predictions satisfying
the commitment criterion into parallel commitments.

We evaluate dLoop against Fast-dLLM, dParallel, and d3LLM on GSM8K, MATH,
MBPP, and HumanEval, using task accuracy and tokens per forward pass (TPF)
with the computation of extra loops included. dLoop achieves the highest or
tied-highest accuracy on every benchmark, and no compared method surpasses it
in both accuracy and TPF. Unlike dParallel and d3LLM, which push predictions
to become confident earlier through a certainty-forcing loss, dLoop keeps the
standard masked-token cross-entropy and lets confidence rise through
additional computation before commitment. This demonstrates that
pre-commitment latent computation offers an effective acceleration route
distinct from confidence training, cache optimization, and commitment
scheduling.

We further analyze how extra loops affect internal states and subsequent
generation. Extra loops primarily act on slow and unstable generation
trajectories. Hidden-state and attention analyses show that uncommitted
representations are reorganized and locally reselect salient attention
connections during the additional computation. Source-rollback interventions
further identify their joint update as the main source of changes in token
candidates and commitment decisions. The unchanged Coda then maps these
refined representations to new token distributions; newly committed tokens
become context for later denoising steps and alter the subsequent trajectory.
These analyses connect recurrent latent computation and distributed
refinement of uncommitted representations to token commitment and complete
generation.

Our contributions are threefold:
\begin{enumerate}
    \item We introduce dLoop, which adds recurrent latent computation for
    immature predictions before irreversible token commitment without
    creating a new external denoising state.

    \item We show across four reasoning and code-generation benchmarks that
    dLoop achieves the highest or tied-highest accuracy on every benchmark and
    a highly competitive accuracy--parallelism trade-off against Fast-dLLM,
    dParallel, and d3LLM.

    \item We provide a mechanistic analysis showing how extra loops reorganize
    uncommitted representations, change token candidates and commitment
    decisions, and ultimately affect subsequent generation.
\end{enumerate}
\section{Related Work}
\begin{figure*}[!t]
    \centering
    \includegraphics[width=0.92\textwidth]
    {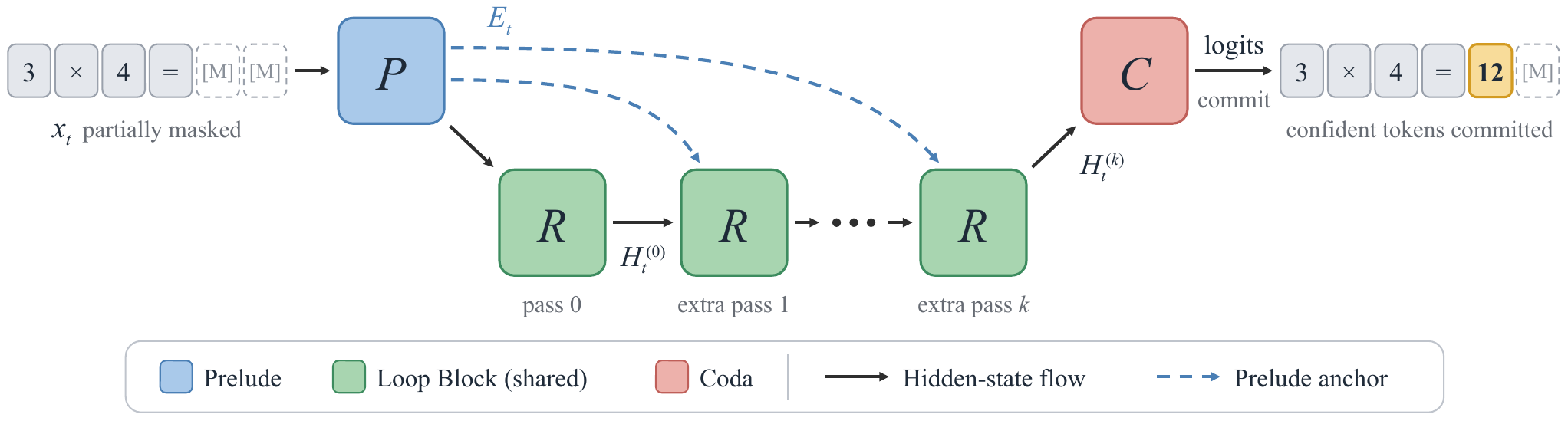}
    \caption{Overview of dLoop. Here, \(X_t\) denotes the partially
    masked input sequence at denoising step \(t\), and \(E_t\) is the
    corresponding Prelude output, which remains fixed as an input anchor
    throughout the extra passes. \(H_t^{(0)}\) denotes the hidden state
    produced by the initial recurrent pass, whereas \(H_t^{(k)}\) denotes
    the refined hidden state after \(k\) extra passes through the
    parameter-shared recurrent block. The Coda maps \(H_t^{(k)}\) to
    output logits.}
    \label{fig:dlm_loop_architecture}
\end{figure*}
\paragraph{Efficient Decoding for Diffusion Language Models.}
Diffusion language models replace a fixed left-to-right factorization with iterative prediction over masked positions, enabling flexible generation orders and multi-token commitment \citep{nie2025llada}.
Their practical speed is nevertheless governed by an accuracy--parallelism trade-off: aggressive commitment reduces denoising steps but can turn immature predictions into irreversible context.
Existing work improves this trade-off by changing either the training objective or the decoding policy.
dParallel uses certainty-forcing distillation to accelerate certainty convergence across masked positions \citep{chen2025dparallel}, whereas d3LLM combines pseudo-trajectory distillation with entropy-guided multi-block decoding and cache refresh \citep{qian2026d3llm}.
Training-free approaches instead modify inference: Fast-dLLM combines confidence-aware parallel decoding with approximate KV caching \citep{wu2025fastdllm}, while Deferred Commitment Decoding uses a confidence-aware sliding window to postpone uncertain decisions \citep{shu2026dcd}.
Together, these methods accelerate diffusion decoding either by making more tokens ready for earlier commitment or by reducing the cost of repeated denoising.

\paragraph{Recurrent and Looped Language Models.}
Recurrent and looped Transformers repeatedly apply a parameter-shared block, allowing test-time computation to vary without duplicating its parameters.
Recurrent-depth models establish this mechanism as a latent test-time-compute axis \citep{geiping2025recurrent}, while Retrofitted Recurrence develops a training procedure for converting pretrained fixed-depth language models into recurrent ones \citep{mcleish2025retrofitted}.
Parcae studies the stability and scaling behavior of looped residual dynamics and introduces a constrained input-injection mechanism \citep{prairie2026parcae}.
Other work explores how loop depth should be allocated: LoopFormer learns representations across variable-length recurrent trajectories \citep{jeddi2026loopformer}, whereas Think-at-Hard selectively applies additional latent iterations to difficult token predictions \citep{fu2025thinkathard}.
Mechanistic analysis of looped autoregressive models identifies cyclic fixed-point behavior, attention stabilization, and repeated stages of inference across recurrent depth \citep{blayney2026mechanistic}.
Most directly for diffusion language modeling, LoopMDM applies selective looping to early--middle layers of masked diffusion models and demonstrates training- and inference-depth scaling, adaptive stopping, and increased interaction among masked positions \citep{lee2026loopmdm}.

\paragraph{Latent Computation and Mechanistic Analysis.}
Reasoning with Latent Tokens shows that still-masked positions are not computationally empty: their evolving predictions can support decisions elsewhere, and varying their number produces a speed--quality trade-off \citep{he2026latent}.
Latent Refinement Decoding approaches the same broad issue from decoding, treating premature commitment as information loss and retaining distributional beliefs across denoising steps \citep{zhu2025lrd}.
Together, these studies motivate viewing uncommitted positions as an evolving latent workspace rather than as empty placeholders awaiting prediction.

\paragraph{Relation to Our Work.}
Prior work separately establishes efficient token commitment, recurrent depth as an internal compute axis, and the latent representations carried by uncommitted positions.
dLoop connects these directions by placing trained recurrent computation before token commitment and tracing how updates to the uncommitted latent workspace change confidence, commitment decisions, and subsequent denoising.
\section{Method}

At each denoising step, dLoop refines the current partially masked state
through recurrent computation before the decoder commits tokens. We next
describe its recurrent architecture, multi-depth training, and decoding
procedure.

\subsection{dLoop Recurrent Model}

\paragraph{Model architecture.}
We divide LLaDA \citep{nie2025llada} into three consecutive
functional components: a Prelude that encodes the current partially masked
sequence, a parameter-shared Loop Block that performs repeated computation,
and a Coda that reads the updated representations and produces token
predictions. Inspired by recurrent-depth architectures
\citep{geiping2025recurrent,mcleish2025retrofitted}, we use a consecutive
group of intermediate layers as the repeatable computation block: layers
0--15 form the Prelude, layers 16--21 form the Loop Block, and layers 22--31
form the Coda. Figure~\ref{fig:dlm_loop_architecture} illustrates this architecture and
the recurrent computation performed within each denoising step.

For a partially masked sequence \(x_t\) at denoising step \(t\), the Prelude
is evaluated once:
\[
    E_t=F_{\mathrm{pre}}(x_t).
\]
The first application of the Loop Block follows the original feed-forward
path:
\[
    H_t^{(0)}=F_{\mathrm{loop}}(E_t).
\]
We set a maximum extra-loop depth \(K\), and let
\(k\in\{0,\ldots,K\}\) denote the number of extra loops used for the current
example. Here \(k=0\) executes only the original path, whereas \(k>0\) adds
\(k\) further applications of the same Loop Block. The Loop Block therefore
executes \(k+1\) times in total.

At extra pass \(j=1,\ldots,k\), a fusion function \(I_\phi\) first combines
the preceding Loop Block output with the fixed Prelude representation:
\[
    X_t^{(j)}
    =I_\phi\!\left(H_t^{(j-1)},E_t\right).
\]
The shared Loop Block is then executed again:
\[
    H_t^{(j)}
    =F_{\mathrm{loop}}\!\left(X_t^{(j)}\right).
\]

Finally, the Coda processes the last Loop-Block state, and the LM head maps
the Coda output to vocabulary logits:
\[
    \ell_t^{(k)}
    =\operatorname{LMHead}\!\left(
      F_{\mathrm{coda}}(H_t^{(k)})
    \right).
\]
These logits define the token distribution
\(p_\theta(\cdot\mid x_t,k)\) used for training and decoding.
The input \(x_t\) and its Prelude representation \(E_t\) remain fixed
throughout all extra passes.

\paragraph{Prelude-anchored extra-loop update.}
Each extra pass must use two sources of information: the preceding Loop Block
output \(H\), which preserves the computation already performed, and the
fixed Prelude representation \(E\), which anchors the recurrence to the
current input. Reusing only \(H\) may repeatedly amplify the residual state,
whereas restarting only from \(E\) would discard the preceding computation.
We therefore adopt the stable input-injection parameterization of Parcae
\citep{prairie2026parcae} to retain prior computation while reinjecting the
Prelude representation.

Let \(H,E\in\mathbb R^{L\times d}\), where \(L\) is the sequence length and
\(d\) is the hidden dimension. The fusion has learnable parameters
\(\phi=\{a,b,B\}\), where \(a,b\in\mathbb R^d\) are feature-wise vectors and
\(B\in\mathbb R^{d\times d}\) is an input-projection matrix. These parameters
define an update step \(\delta\in\mathbb R^d\) and a state-retention matrix
\(D\in\mathbb R^{d\times d}\). The symbol \(\odot\) denotes element-wise
multiplication.

The fused input to each extra pass is
\[
    I_\phi(H,E)
    =HD+\delta\odot(EB).
\]
The first term carries forward the computation from the preceding pass,
whereas the second reinjects the fixed Prelude representation. We next define
the retention matrix \(D\), update step \(\delta\), and projection matrix
\(B\).

We transform the learnable vector \(b\) element-wise into a positive update
step:
\[
    \delta=\log\!\left(1+\exp(b)\right)>0.
\]
Another learnable vector \(a\) defines a feature-wise state-retention matrix:
\[
    D=\operatorname{Diag}\!\left(
      \exp[-\delta\odot\exp(a)]
    \right).
\]
Because \(\delta_i>0\) and \(\exp(a_i)>0\), every diagonal element
\(D_{ii}=\exp[-\delta_i\exp(a_i)]\) lies strictly between zero and one.
Consequently,
\[
    \rho_{\mathrm{spec}}(D)
    =\max_i |D_{ii}|<1.
\]
This constraint controls how much of each hidden feature is retained and
prevents the direct carry path of \(H\) from being amplified across extra
passes. All extra passes within the same forward computation share the same
\(D\).

The matrix \(B\) is not constrained to be diagonal and learns how to project
the Prelude representation back into the recurrent state. Thus, \(D\) and
\(B\) play distinct roles: \(D\) controls stable state propagation, whereas
\(B\) transforms the anchoring information.

\subsection{Multi-Depth SFT}

During training, we assign different examples extra-loop depths
\(k\in\{0,\ldots,K\}\) and optimize standard masked-token cross-entropy on
partially masked inputs:
\[
    \mathcal L_{\mathrm{SFT}}
    =-\frac{1}{|\mathcal M|}
      \sum_{i\in\mathcal M}
      \log p_\theta(y_i\mid x,k).
\]
Here \(x\) is the partially masked input, \(\mathcal M\) is its set of masked
positions, and \(y_i\) is the target token at position \(i\). Training across
different extra-loop depths enables the parameter-shared Loop Block to operate
under different internal-compute budgets.

\subsection{Decode}

The recurrent model and decoder play complementary roles: extra loops refine
the predictive distributions at uncommitted positions, while the decoder
decides which predictions are ready to become visible context. We use a
confidence-aware expanding frontier that restricts commitment to a moving
local region while allowing that region to expand so that it retains enough
uncommitted candidates.

At denoising step \(t\), let \(\mathcal U_t\) denote the response positions
that remain masked. Starting from the leftmost position in \(\mathcal U_t\),
the decoder first considers a nominal span of \(L_{\mathrm{win}}\) response
positions. If this span contains fewer than \(M\) uncommitted positions, its
right boundary is extended until it includes the \(M\)-th uncommitted
position or reaches the end of the response. We denote the resulting
candidate set by
\[
    \mathcal W_t\subseteq\mathcal U_t.
\]
The span \(L_{\mathrm{win}}\) keeps commitment near the current generation
frontier, while \(M\) prevents the candidate pool from becoming too small as
earlier positions are committed.

For each candidate position \(i\in\mathcal W_t\), confidence is the
probability of its most likely token:
\[
    C_t(i)=\max_y p_\theta(x_i=y\mid x_t,k),
\]
where \(p_\theta(x_i=y\mid x_t,k)\) is the probability of assigning token
\(y\) to position \(i\), conditioned on the current partially masked sequence
\(x_t\) and \(k\) extra loops. Every candidate satisfying
\[
    C_t(i)\geq\tau
\]
is selected.

To ensure that generation finishes within the denoising budget, let \(s_t\)
be the number of remaining steps. The decoder requires at least
\[
    m_t
    =\left\lceil\frac{|\mathcal U_t|}{s_t}\right\rceil
\]
commitments, subject to the number of candidates available. If fewer than
\(m_t\) candidates exceed \(\tau\), the decoder fills the remainder with the
highest-confidence candidates in \(\mathcal W_t\).

The confidence threshold turns predictions that mature through internal
computation into additional parallel commitments, while the minimum-progress
rule prevents stalled generation. Once selected, these tokens are written
into the sequence and become fixed context for subsequent denoising steps.
\section{Experiments and Analysis}
\subsection{Experiments}

\paragraph{Models and Benchmarks.}
All experiments build on LLaDA. dLoop uses the recurrent architecture described in
the Method section, with layers 16--21 serving as the parameter-shared Loop
Block. At inference time, we use the confidence-aware decoder described in
the Method section to commit mature predictions within a moving local window.
We evaluate mathematical reasoning on GSM8K and MATH, and code generation on
MBPP and HumanEval. We compare with native LLaDA, Fast-dLLM
\citep{wu2025fastdllm}, dParallel \citep{chen2025dparallel}, and d3LLM
\citep{qian2026d3llm}. Training details are given in the appendix.

\paragraph{Evaluation Metrics.}
We evaluate each method using task accuracy and tokens per forward pass
(TPF). Accuracy measures generation quality, while TPF measures the average
number of output tokens generated by one full-model denoising forward pass.
Let \(N_{\mathrm{tok}}\) be the total number of generated output tokens and
\(N_{\mathrm{fwd}}\) the number of full-model forward passes required for
generation. Raw TPF is
\[
\mathrm{TPF}_{\mathrm{raw}}
=
\frac{N_{\mathrm{tok}}}{N_{\mathrm{fwd}}}.
\]
A higher TPF indicates that the model commits more tokens per forward pass
and therefore achieves greater decoding parallelism.

Raw TPF counts only external denoising passes and ignores the partial-depth
computation introduced by extra loops. We therefore convert each extra loop
into its equivalent fraction of a full forward pass. Because dLoop repeats
six layers in a 32-layer backbone, a configuration with \(k\) extra loops is
assigned
\[
\mathrm{TPF}_{\mathrm{adj}}
=
\frac{\mathrm{TPF}_{\mathrm{raw}}}{1+6k/32}.
\]
This metric measures generated tokens per equivalent full-model forward pass.

\subsubsection{Main Results}

\begin{table}[!t]
\centering
{\small
\setlength{\tabcolsep}{8pt}
\renewcommand{\arraystretch}{1.15}

\begin{tabular}{llcc}
\toprule
\textbf{Benchmark} & \textbf{Method} & \textbf{TPF} \(\uparrow\) &
\textbf{Acc.\ (\%)} \(\uparrow\) \\
\midrule
    & LLaDA          & \(1.00\)          & \(72.6\) \\
    & Fast-dLLM      & \(2.77\)          & \(75.1\) \\
    & dParallel      & \(5.11\)          & \(72.9\) \\
    & d3LLM          & \besttpf{9.06}    & \(72.9\) \\
\rowcolor{OursRow} \cellcolor{white}\multirow{-5}{*}{GSM8K}
    & \textbf{dLoop} & \(6.76\)          & \bestacc{76.3} \\
\midrule
    & LLaDA          & \(1.00\)          & \bestacc{32.2} \\
    & Fast-dLLM      & \(1.94\)          & \(31.2\) \\
    & dParallel      & \(3.15\)          & \(30.0\) \\
    & d3LLM          & \besttpf{5.95}    & \(30.5\) \\
\rowcolor{OursRow} \cellcolor{white}\multirow{-5}{*}{MATH}
    & \textbf{dLoop} & \(4.59\)          & \bestacc{32.2} \\
\midrule
    & LLaDA          & \(1.00\)          & \(41.7\) \\
    & Fast-dLLM      & \(2.12\)          & \(38.0\) \\
    & dParallel      & \(2.26\)          & \(40.6\) \\
    & d3LLM          & \besttpf{4.59}    & \(41.4\) \\
\rowcolor{OursRow} \cellcolor{white}\multirow{-5}{*}{MBPP}
    & \textbf{dLoop} & \(3.31\)          & \bestacc{42.2} \\
\midrule
    & LLaDA          & \(1.00\)          & \(38.3\) \\
    & Fast-dLLM      & \(2.47\)          & \(37.8\) \\
    & dParallel      & \(4.96\)          & \bestacc{39.6} \\
    & d3LLM          & \besttpf{5.34}    & \(37.8\) \\
\rowcolor{OursRow} \cellcolor{white}\multirow{-5}{*}{HumanEval}
    & \textbf{dLoop} & \(4.08\)          & \bestacc{39.6} \\
\bottomrule
\end{tabular}
}
\caption{Accuracy and tokens per forward pass (TPF) for LLaDA-based
models. dLoop uses adjusted TPF. \textcolor{BestBlue}{\textbf{Blue}}
marks the best TPF and \textcolor{BestRed}{\textbf{red}} the best
accuracy on each benchmark; rows shaded in
\colorbox{OursRow}{\strut light blue} are our method.}
\label{tab:main_results}
\end{table}

As shown in Table~\ref{tab:main_results}, dLoop achieves the highest
accuracy on GSM8K (\(76.3\%\)) and MBPP (\(42.2\%\)), and ties for the
highest accuracy on MATH (\(32.2\%\), with native LLaDA) and HumanEval
(\(39.6\%\), with dParallel). It matches or exceeds the accuracy of native
LLaDA on every benchmark while decoding with substantially greater
parallelism. Compared with Fast-dLLM, dLoop achieves both higher accuracy and
higher adjusted TPF on all four benchmarks; compared with dParallel, it does
so on GSM8K, MATH, and MBPP, while on HumanEval the two methods reach the
same accuracy and dParallel attains higher TPF. Compared with d3LLM, the
advantage splits: d3LLM attains higher TPF on all four benchmarks, whereas
dLoop achieves higher accuracy on all four.

This split between accuracy and TPF reflects the trade-off between accuracy
and parallelism in decoding acceleration, and it stems from how the two kinds
of methods raise prediction confidence. dParallel and d3LLM work through the
loss: a certainty-forcing loss directly pushes predictions to become
confident earlier, so that more tokens can be committed per step; d3LLM
builds further optimizations on dParallel while keeping the same core idea.
dLoop instead works through the model structure: its loss remains the
standard masked-token cross-entropy, with no confidence-specific term in the
training objective, and it adds a recurrent loop block so that predictions
receive additional computation before commitment. Confidence then rises
spontaneously through this additional latent computation, allowing
predictions to mature faster and thereby also accelerating decoding. The two
routes therefore emphasize different goals. Forcing confidence through the
loss favors parallelism, and together with the further optimizations of
d3LLM, this gives d3LLM the highest TPF. Letting confidence mature through
the structure favors accuracy, while the extra computation of the loops is
fully counted in adjusted TPF, so dLoop does not reach the TPF of d3LLM.
Taken together, on every benchmark in Table~\ref{tab:main_results}, no other
method achieves both higher accuracy and higher TPF than dLoop. dLoop is
thus a highly competitive choice on the accuracy--parallelism trade-off, and
it represents a route to DLM acceleration distinct from dParallel and d3LLM.
\subsection{Analysis}

The main results show that dLoop improves the
accuracy--parallelism trade-off. We next analyze why the additional
pre-commitment computation helps. Our analysis first identifies the
behavioral failure addressed by an extra loop, then uses single-step
interventions to locate the responsible hidden-state update, and finally
tests whether this local effect accumulates during complete decoding.

\subsubsection{Extra Loops Help When Predictions Are Unstable}

We take the first 300 examples, in order, from the validation data used
during model development. Overall, adding one extra loop increases accuracy from
\(88.0\%\) to \(93.0\%\) and reduces mean decoding steps from \(23.85\) to
\(22.39\). The gain is not distributed uniformly. Without an extra loop, a
subset of difficult examples requires many denoising steps to finish
generation. We define examples requiring more than 32 steps as long
trajectories and identify 43 such cases. On this subset, one extra loop
raises accuracy from \(69.8\%\) to \(88.4\%\) and reduces mean decoding
steps from \(38.4\) to \(31.9\).

These long trajectories contain unstable predictions inside the current
commitment window. If the same uncommitted position remains in the window
for two consecutive denoising steps but changes its top-1 token, we count
one \emph{flip}. Summed over all window positions and denoising steps, the 43
long trajectories contain \(338.7\) flips per example. Adding one extra loop
reduces this average to \(262.0\).

The repeated flips arise when the top-1 and top-2 probabilities are close.
A small change in the partially generated context can then reverse their
order. If too few positions reach the confidence threshold, the
minimum-progress rule may irreversibly commit a plausible but immature
candidate. The extra loop provides additional computation before
commitment, making the leading candidate more likely to separate from its
competitor and reach the decoding threshold. We next examine which internal
update gives rise to this stabilizing effect.

\subsubsection{Which Part of the Extra-Loop Update Matters?}

Prior work shows that still-masked positions in a diffusion language model
can carry intermediate candidate information rather than acting as empty
placeholders \citep{he2026latent}. Because an extra loop recomputes the
entire partially masked sequence, it updates these uncommitted
representations together with the prompt and previously committed context.
This raises a key question: which source of the extra-loop update drives the
change in the current target prediction?

\begin{figure}[!t]
    \centering
    \includegraphics[width=0.80\columnwidth]
    {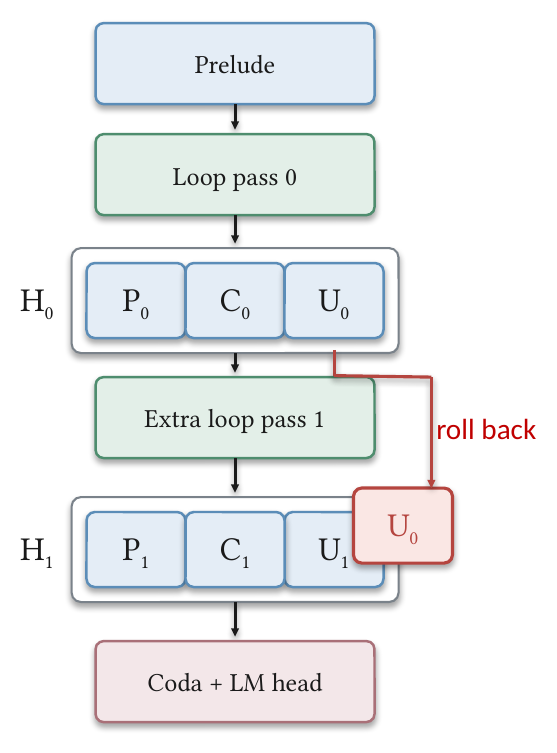}
    \caption{Single-step source rollback. \(H_0=[P_0,C_0,U_0]\) and
    \(H_1=[P_1,C_1,U_1]\) denote the hidden states produced before and
    after one extra loop, respectively. \(P\), \(C\), and \(U\) denote
    the hidden states at prompt-token positions, previously committed-token
    positions, and currently uncommitted positions. Subscripts \(0\) and
    \(1\) indicate states before and after the extra loop. The intervention
    constructs \([P_1,C_1,U_0]\) by rolling back only the uncommitted
    component from \(U_1\) to \(U_0\), while retaining \(P_1\), \(C_1\),
    and the same Coda and language-model head.}
    \label{fig:source_rollback}
\end{figure}

Let \(H_t^{(0)}\) denote the Loop-Block output at denoising step \(t\) when
no extra loop is used, and let \(H_t^{(1)}\) denote the output obtained from
the same partially masked sequence after one additional pass. Both are
measured before the Coda, and no token is committed between them. The extra
loop therefore produces
\[
    H_t^{(1)}=H_t^{(0)}+\Delta H_t.
\]
Each row of \(H_t\) corresponds to one sequence position. We divide the
update by source into prompt tokens (\(P\)), previously committed tokens
(\(C\)), and uncommitted positions (\(U\)):
\[
    \Delta H_t=
    [\Delta H_t^{P},\Delta H_t^{C},\Delta H_t^{U}].
\]
The unchanged Coda and output head map these representations to logits,
margins, and commitment decisions; the intervention below changes only
which component of the extra-loop update is retained. As shown in
Figure~\ref{fig:source_rollback}, we start from the complete extra-loop
state, roll back one source to its pre-loop representation, and observe how
the target prediction changes.

\subsubsection{Single-Step Source Rollback}

At a fixed denoising step, the complete extra-loop branch uses
\([P_1,C_1,U_1]\). To remove one source update, we replace only that source
with its no-extra-loop state, yielding \([P_0,C_1,U_1]\),
\([P_1,C_0,U_1]\), or \([P_1,C_1,U_0]\). All remaining representations and
the readout network are unchanged.

For each branch, we measure the target's top-1/top-2 logit margin,
\[
    m=z_{\mathrm{top1}}-z_{\mathrm{top2}},
\]
and whether its top-1 candidate changes. The top two candidates are
identified independently in each branch and may therefore change after
rollback. A larger margin indicates that the leading candidate is more clearly
separated from its closest competitor and is thus more stable.

Table~\ref{tab:source_rollback} reports the average results on the same
validation data as above. Its margin decrease compares the complete extra-loop branch with
the rollback branch:
\[
    \Delta m=m_{\mathrm{full}}-m_{\mathrm{rollback}}.
\]
A larger positive value means that rollback brings the top two candidates
closer together, indicating that the reverted update contributed more to
separating the candidates and stabilizing the current prediction. A value
near zero indicates little average change.

\begin{table}[t]
\centering
{\small
\begin{tabular}{@{}lcc@{}}
\toprule
Reverted update & Margin decrease & Top-1 changed \\
\midrule
Prompt \(\Delta H^P\) & \(0.043\) & \(19.4\%\) \\
Committed \(\Delta H^C\) & \(-0.004\) & \(10.4\%\) \\
Uncommitted \(\Delta H^U\) & \(\mathbf{0.130}\) & \(\mathbf{47.8\%}\) \\
\bottomrule
\end{tabular}
}
\caption{Single-step rollback of different components of
\(\Delta H_t\).}
\label{tab:source_rollback}
\end{table}

Reverting \(\Delta H^U\) produces the largest change on both measures: the
target margin decreases by \(0.130\), and \(47.8\%\) of the targets select a
different top-1 candidate. Reverting prompt or committed updates has a much
smaller effect. Thus, the uncommitted component of \(\Delta H_t\) is the
main source of the immediate change in the current prediction. Due to
space limits, additional results are given in the appendix:
Table~\ref{tab:rollback_tasks} shows the same pattern over all masked
targets on GSM8K, MATH, MBPP, and HumanEval.

\subsubsection{From Single-Step Changes to Their Accumulated Effect}

The preceding experiment measures only the immediate change in the target
margin and top-1 candidate at one fixed denoising step. It does not show
whether repeatedly preserving the same update source affects the efficiency
or accuracy of the complete generation.

We therefore perform full decoding on the first 700 GSM8K test examples and
apply source rollback at every denoising step. We compare an
uncommitted-only condition, \([P_0,C_0,U_1]\), which retains only the
extra-loop update at uncommitted positions, with a
prompt-and-committed-only condition, \([P_1,C_1,U_0]\). We also report the full extra loop and full
rollback as references. The model, examples, and decoding configuration
are identical; the conditions differ only
in which part of the extra-loop update is retained.

After each rollback, the controlled state passes through the same Coda and
output head, and the decoder commits tokens normally. The resulting
partially generated sequence is then fed back as the input to the next
denoising step, where the loop, rollback, and commitment procedure is
repeated until generation is complete. We finally compare answer accuracy
and the total number of denoising steps.

Table~\ref{tab:persistent_source} reports the complete-decoding results.

\begin{table}[t]
\centering
{\small
\begin{tabular}{@{}lcc@{}}
\toprule
Condition & Accuracy & Mean steps \\
\midrule
Full extra loop \([P_1,C_1,U_1]\) & \(74.43\%\) & \(24.62\) \\
Updated \(U\) only \([P_0,C_0,U_1]\) & \(73.14\%\) & \(24.96\) \\
Updated \(P/C\) only \([P_1,C_1,U_0]\) & \(71.43\%\) & \(26.10\) \\
Full rollback \([P_0,C_0,U_0]\) & \(70.86\%\) & \(26.81\) \\
\bottomrule
\end{tabular}
}
\caption{Persistent source control during complete decoding on the first
700 GSM8K test examples. Full rollback is identical to decoding without an extra loop.}
\label{tab:persistent_source}
\end{table}

The uncommitted-only condition achieves higher accuracy than the
prompt-and-committed-only condition while using \(1.14\) fewer denoising
steps on average. This result shows that the uncommitted update remains
useful throughout generation and converts the single-step margin advantage
into more efficient complete decoding.

\subsubsection{How Does the Extra Loop Modify Uncommitted Representations?}

The single-step and complete-decoding experiments identify the uncommitted
component as the more influential part of \(\Delta H_t\). We next examine
what changes at these positions. For position \(i\), we measure the relative
hidden-state displacement
\[
    r_i=
    \frac{\lVert H_{t,i}^{(1)}-H_{t,i}^{(0)}\rVert_2}
         {\lVert H_{t,i}^{(0)}\rVert_2}.
\]

Table~\ref{tab:hidden_change} summarizes the relative hidden-state
displacement across position groups.

\begin{table}[t]
\centering
{\small
\begin{tabular}{@{}lc@{}}
\toprule
Position group & Relative displacement \(r\) \\
\midrule
Uncommitted \(U\) & \(\mathbf{0.275}\) \\
Committed \(C\) & \(0.233\) \\
\bottomrule
\end{tabular}
}
\caption{Hidden-state changes produced by one extra loop.}
\label{tab:hidden_change}
\end{table}

Uncommitted positions change more relative to their original states than
committed positions, showing that the extra pass more strongly rewrites
representations at positions that have not yet been committed.

A plausible interpretation is that committed positions already correspond
to visible tokens and retain relatively stable semantic content, whereas
uncommitted positions are still represented by \texttt{[MASK]} at the
input. Their position-specific candidate information is carried more
strongly by context-dependent hidden representations and can therefore be
reorganized through additional information exchange in the loop block.

We further inspect whether the attention computation changes while forming
these updates. For queries inside the commitment window, we compare the
top-\(k\) key positions before and after the extra pass across attention
heads and loop layers. The retention rate is the overlap between the two
top-\(k\) sets.

Table~\ref{tab:attention_topk} reports the resulting retention rates.

\begin{table}[t]
\centering
{\small
\begin{tabular}{@{}lc@{}}
\toprule
Attention scope & Top-\(k\) retention \\
\midrule
Top-1 & \(70.7\%\) \\
Top-5 & \(73.4\%\) \\
Top-10 & \(74.8\%\) \\
Top-20 & \(76.5\%\) \\
Top-40 & \(78.6\%\) \\
\bottomrule
\end{tabular}
}
\caption{Overlap of the most-attended key positions before and after one
extra loop.}
\label{tab:attention_topk}
\end{table}

The top-1 key retention is \(70.7\%\), and the retention remains only
\(78.6\%\) when the scope is expanded to the top 40 keys. These results show
that the extra loop locally reselects which context positions receive the
strongest attention, allowing this attention reorganization to participate
in forming the new uncommitted representations.

Together, the results support the following mechanism. An extra loop
recomputes the partially masked sequence and produces
\(\Delta H_t=[\Delta H_t^P,\Delta H_t^C,\Delta H_t^U]\). The uncommitted
representations undergo larger relative changes, while local attention
patterns reselect part of the context used to form these states. The
unchanged Coda and output head convert the resulting representations into
new logits and commitment decisions. Mature candidates can be further
stabilized and committed, whereas uncertain candidates can be revised and
remain uncommitted. Newly committed tokens then become visible context for
the next denoising step. Repeating this process changes the complete
generation trajectory and improves decoding efficiency.

From the perspective of latent reasoning, these results suggest that
uncommitted positions are not independent empty placeholders but jointly
form a distributed latent workspace. During an extra loop, these positions
continue to exchange and reintegrate information with other positions
through attention while updating the candidate representations they carry.
Source rollback further shows that the joint update of these uncommitted
latent tokens is read by the Coda and contributes most strongly to the
current candidate margin and token choice. We therefore interpret
dLoop as performing latent refinement before irreversible token
commitment: the model continues to compute and exchange information among
uncommitted representations before converting a maturing internal decision
into a formal token.
\section{Conclusion}

We introduced dLoop, which uses recurrent latent computation to refine
immature predictions before irreversible token commitment. By repeatedly
executing a parameter-shared intermediate block on the same partially masked
state, dLoop extends internal processing without advancing to a new
denoising step. Across four reasoning and code-generation benchmarks, it
achieves the highest or tied-highest accuracy on every benchmark, and no
compared method, including Fast-dLLM, dParallel, and d3LLM, surpasses it in
both accuracy and TPF. Unlike methods that force confidence through the
training loss, dLoop lets confidence rise through additional computation
before commitment, offering a distinct route to DLM acceleration. Our
analyses show that extra loops perform distributed latent refinement by
reorganizing uncommitted representations and locally reselecting salient
context. These changes alter the leading candidate and its separation from
alternatives, affect commitment decisions, and propagate through subsequent
denoising, connecting pre-commitment latent refinement to the observed
accuracy--parallelism improvements.

\bibliography{references}
\appendix

\section{Appendix}
\label{app:impl}

\subsection{Training Details}
\label{app:training}

We initialize dLoop from LLaDA-8B-Instruct \citep{nie2025llada} and
fine-tune all parameters on the training data released by d3LLM
\citep{qian2026d3llm}.
During training, the number of extra loops is at most \(K=3\).
The number of extra loops used for each training example is assigned as
follows: each parameter update uses 64 examples, and the depths
\(k\in\{0,1,2,3\}\) are assigned to these examples in a round-robin manner,
so each depth receives 16 examples and all four depths are trained in every
update.
The fusion parameters are initialized so that \(I_\phi(H,E)=0.5H+0.5E\).

\subsection{Choice of the Loop Block}
\label{app:layers}

We chose the loop block by training 11 contiguous blocks of 4, 6, or 8
layers with the same data and recipe, and evaluating each block on the first
400 GSM8K test problems with 1, 2, and 3 extra loops
(Table~\ref{tab:layer_sweep}).
Blocks that end at or before layer 15 perform worst.
Layers 16--21 achieve the highest average accuracy over the three depths
(79.0\%, compared with 77.6\% for the runner-up, layers 12--19) as well as
the highest single result (80.5\% at \(k=2\), compared with 78.8\% for the
runner-up, layers 16--23).
Each individual result has a standard error of about 2 points.
We use this block for dLoop.

\begin{table}[h]
\centering
{\small
\begin{tabular}{@{}lccc@{}}
\toprule
Loop block & Layers & Avg.\ (\(k=1\)--3) & Best (\(k\)) \\
\midrule
6--9   & 4 & 74.83 & 76.25 (3) \\
10--13 & 4 & 74.00 & 75.50 (3) \\
14--17 & 4 & 75.42 & 77.75 (3) \\
18--21 & 4 & 76.67 & 77.75 (2) \\
22--25 & 4 & 76.17 & 77.00 (2) \\
10--15 & 6 & 74.00 & 74.75 (3) \\
13--18 & 6 & 75.17 & 76.75 (3) \\
16--21 & 6 & \textbf{79.00} & \textbf{80.50} (2) \\
8--15  & 8 & 74.08 & 74.75 (3) \\
12--19 & 8 & 77.58 & 78.00 (3) \\
16--23 & 8 & 76.92 & 78.75 (2) \\
\bottomrule
\end{tabular}
}
\caption{GSM8K accuracy (\%) of different loop blocks on the first 400 test
problems: the average over 1--3 extra loops and the best result among them.}
\label{tab:layer_sweep}
\end{table}

\subsection{Number of Extra Loops}
\label{app:depth}

\paragraph{Depths within training.}
The model is trained with 0--3 extra loops.
The first four columns of Table~\ref{tab:depth} show the results within this
range: accuracy peaks at \(k=2\) and already decreases slightly at \(k=3\)
(GSM8K: 79.15 to 78.17; MATH: 27.37 to 26.17), whereas adjusted TPF keeps
decreasing.
There are two possible reasons.
First, all four depths share the same loop-block parameters and must be
accommodated simultaneously during training; they may pull these parameters
in different directions, so the deepest setting is not necessarily fitted
best.
Second, more loops may push the representations further away from those the
Coda is accustomed to reading.
Training with more extra loops therefore yields limited benefit, and 0--3
extra loops are sufficient.

\paragraph{Multi-depth training.}
The following comparison also shows that different depths affect one
another.
We also train models with a single fixed depth: training only with \(k=2\)
lowers accuracy without an extra loop (71.7\% vs.\ 75.0\%).
Training with several depths keeps every depth usable in one model, so the
number of extra loops can be adjusted at inference.

\paragraph{Extrapolation to unseen depths.}
At inference, we further increase \(k\) to 4 and 5, which are not used during
training, to examine how the model behaves at unseen depths.
Accuracy does not collapse and remains close to that at \(k=3\) (GSM8K: 78.09
and 78.39; MATH: 26.47 and 26.77), showing that the loop remains stable at
depths not seen during training.
However, accuracy does not improve further, while adjusted TPF continues to
decrease (to 5.41 and 4.93 on GSM8K).
We therefore do not train with more extra loops.

\begin{table}[h]
\centering
{\footnotesize
\setlength{\tabcolsep}{3pt}
\begin{tabular}{@{}lcccccc@{}}
\toprule
\(k\) & 0 & 1 & 2 & 3 & \(4^{\dagger}\) & \(5^{\dagger}\) \\
\midrule
GSM8K Acc.       & 74.98 & 78.01 & \textbf{79.15} & 78.17 & 78.09 & 78.39 \\
GSM8K Adj.\ TPF  & 8.77  & 7.71  & 6.77  & 6.04  & 5.41  & 4.93  \\
MATH Acc.        & 25.37 & 27.17 & \textbf{27.37} & 26.17 & 26.47 & 26.77 \\
MATH Adj.\ TPF   & 5.45  & 4.75  & 4.14  & 3.64  & 3.29  & 2.98  \\
\bottomrule
\end{tabular}
}
\caption{Accuracy (\%) and adjusted TPF for different numbers of extra
loops. \(^{\dagger}\)Not used during training.}
\label{tab:depth}
\end{table}

\subsection{Decoding Settings}
\label{app:decoding}

dLoop generates at most 256 tokens with block length 16 and stops at the
first EOS token.
Table~\ref{tab:threshold} varies the confidence threshold \(\tau\) with one
extra loop.

\begin{table}[h]
\centering
{\footnotesize
\setlength{\tabcolsep}{3.5pt}
\begin{tabular}{@{}lcccccc@{}}
\toprule
\(\tau\) & 0.55 & 0.65 & 0.725 & 0.80 & 0.875 & 0.95 \\
\midrule
GSM8K Acc.   & 73.9 & 78.3 & 78.0 & 77.8 & 78.5 & 78.5 \\
GSM8K steps  & 18.9 & 22.1 & 26.3 & 32.3 & 40.7 & 56.2 \\
MATH Acc.    & 22.4 & 27.0 & 27.2 & 28.2 & 30.0 & 29.3 \\
MATH steps   & 26.2 & 31.1 & 37.1 & 44.6 & 54.8 & 72.3 \\
\bottomrule
\end{tabular}
}
\caption{Accuracy (\%) and mean decoding steps for different confidence
thresholds.}
\label{tab:threshold}
\end{table}

\subsection{Adjusted TPF and Wall-Clock Time}
\label{app:tpf}

\paragraph{TPF.}
An autoregressive model generates one token per forward pass, whereas a
diffusion language model can commit several tokens in a single forward pass.
Work on diffusion language models therefore commonly measures decoding
parallelism by TPF: the number of generated tokens divided by the number of
forward passes.

\paragraph{Why adjust TPF.}
This count assumes that every forward pass has the same cost.
The assumption holds for LLaDA, Fast-dLLM, dParallel, and d3LLM, each of
which passes through the 32 layers once per step.
dLoop is different: a step with \(k\) extra loops reruns the 6 loop layers of
the 32-layer backbone \(k\) more times, so it costs more computation than a
step of LLaDA.
Counting such a step as a single forward pass would overstate the speed of
dLoop and make the comparison with other methods unfair.
We therefore use adjusted TPF, which counts a step with \(k\) extra loops as
\((32+6k)/32\) forward passes according to the number of layers executed,
i.e., 1.188, 1.375, and 1.563 for \(k=1,2,3\).
A higher adjusted TPF thus means that the steps saved outweigh the extra
computation spent in each step.

\paragraph{TPF and wall-clock time.}
TPF is not itself a measure of time, but it is closely related to time: the
time to solve a problem equals the number of denoising steps multiplied by
the time per step.
For the same model and implementation, the time per step is roughly constant,
so when the same number of tokens is generated, a higher TPF means fewer steps
and less time.
Each dLoop step becomes slower because of the extra loops, and adjusted TPF
accounts for exactly this cost by converting it according to the number of
layers executed.
As long as the time per step grows in proportion to the number of layers
executed, adjusted TPF agrees with the actual time.

\paragraph{Measured results.}
We measure the time per step on one NVIDIA GH200 GPU with batch size 1.
Relative to a step without an extra loop, a step with 1, 2, and 3 extra loops
takes 1.202, 1.401, and 1.595 times as long (averaged over the four tasks),
which is within 3.4\% of the assumed ratios of 1.188, 1.375, and 1.563 on
every task.
Adjusted TPF therefore matches the actual cost of extra loops.

The same conclusion holds for per-problem latency.
We run dParallel with its official code on the same GPU and dLoop with one
extra loop.
On GSM8K, a dLoop step is slower than a dParallel step (53 vs.\ 46 ms), but
dLoop needs fewer steps (26 vs.\ 37), so a problem takes 1.38 s, compared
with 1.71 s for dParallel.
On MATH, the times are 1.96 s and 2.23 s, respectively.

\subsection{Additional Rollback Analyses}
\label{app:rollback}

We repeat the single-step rollback on MATH, MBPP, and HumanEval
(Table~\ref{tab:rollback_tasks}).
On every task, as on GSM8K, rolling back the other uncommitted positions
changes the top-1 candidate far more often than rolling back the target
itself, the prompt, or the committed positions.

\begin{table}[h]
\centering
{\small
\setlength{\tabcolsep}{4pt}
\begin{tabular}{@{}lcccc@{}}
\toprule
Rolled back & GSM8K & MATH & MBPP & HumanEval \\
\midrule
Prompt            & 6.8  & 8.0  & 7.0  & 5.7  \\
Committed         & 8.0  & 8.5  & 7.3  & 8.6  \\
Target itself     & 4.8  & 5.3  & 4.2  & 4.7  \\
Other uncommitted & \textbf{22.0} & \textbf{20.1} & \textbf{11.9} & \textbf{16.0} \\
\bottomrule
\end{tabular}
}
\caption{Top-1 changes (\%) after single-step rollback over all masked
targets in the commitment window.}
\label{tab:rollback_tasks}
\end{table}

% Keep the official checklist in the project. Uncomment only if the
% AAAI 2027 submission instructions require it to be appended.
% \input{ReproducibilityChecklist.tex}

\end{document}